\documentclass{article}

\usepackage{neurips_data_2024}

\usepackage{hyperref}
\usepackage{array}
\usepackage{multirow}
\usepackage{enumitem}

\usepackage{algorithm}
\usepackage{algorithmicx}
\usepackage{algpseudocode}
\usepackage{graphicx}
\usepackage{subfigure}
\usepackage{wrapfig}
\usepackage{pifont}
\usepackage{amsmath}
\usepackage{amssymb}
\usepackage{mathtools}
\usepackage{amsthm}

\usepackage{comment}
\usepackage{longtable}

\theoremstyle{plain}

\theoremstyle{definition}

\theoremstyle{remark}

\usepackage{makecell}
\usepackage[utf8]{inputenc} 
\usepackage[T1]{fontenc}    
\usepackage{hyperref}       
\usepackage{cleveref}
\usepackage{url}            
\usepackage{booktabs}       
\usepackage{amsfonts}       
\usepackage{nicefrac}       
\usepackage{microtype}      
\usepackage{xcolor}         
\newcommand{\benchshort}{MFE-ETP}

\title{MFM-ETP: An Embodied Task Planning Benchmark for Multi-modal Foundation Models}
\title{MFE-ETP: A Comprehensive Evaluation Benchmark for Multi-modal Foundation Models on Embodied Task Planning}
\author{%
  Min Zhang \textsuperscript{1}\thanks{Equal contribution}, Xian Fu\textsuperscript{1}\footnotemark[1], Jianye Hao\textsuperscript{1} \thanks{Corresponding author: Jianye Hao (jianye.hao@tju.edu.cn)}, Peilong Han\textsuperscript{1}, Hao Zhang\textsuperscript{1},
  \\\textbf{Lei Shi\textsuperscript{1}, Hongyao Tang\textsuperscript{2}, Yan Zheng\textsuperscript{1}} \\
 \textsuperscript{1} College of Intelligence and Computing, Tianjin University\\
 \textsuperscript{2} Montreal Institute of Learning Algorithms (MILA)\\
}

\begin{document}

\maketitle
\begin{abstract}
In recent years, Multi-modal Foundation Models (MFMs) and Embodied Artificial Intelligence (EAI) have been advancing side by side at an unprecedented pace. The integration of the two has garnered significant attention from the AI research community. 
In this work, we attempt to provide an in-depth and comprehensive evaluation of the performance of MFM  s on embodied task planning, aiming to shed light on their capabilities and limitations in this domain. To this end, based on the characteristics of embodied task planning, we first develop a systematic evaluation framework, which encapsulates four crucial capabilities of MFMs: object understanding, spatio-temporal perception, task understanding, and embodied reasoning. 
Following this, we propose a new benchmark, named MFE-ETP, characterized its complex and variable task scenarios, typical yet diverse task types, task instances of varying difficulties, and rich test case types ranging from multiple embodied question answering to embodied task reasoning.
Finally, we offer a simple and easy-to-use automatic evaluation platform that enables the automated testing of multiple MFMs on the proposed benchmark.
Using the benchmark and evaluation platform, we evaluated several state-of-the-art MFMs and found that they significantly lag behind human-level performance. The MFE-ETP is a high-quality, large-scale, and challenging benchmark relevant to real-world tasks.
\end{abstract}

\section{Introduction}
\label{introduction}

The impressive progress of Multi-modal Foundation Models (MFMs) \cite{BLIP-2, MiniGPT-4, InstructBLIP, achiam2023gpt, LLaVA, hu2024minicpm, li2023empowering} has sparked a surge of interest in promoting the application of MFMs in Embodied Artificial Intelligence (EAI).
Currently, the mainstream approaches can be divided into two branches, one of which is to train specific models for low-level
robotic control based on particular robotics data \cite{brohan2022rt,brohan2023rt,li2023vision,shah2023mutex,li2023mastering,driess2023palm,jiang2023vima}. The other is to employ off-the-shelf models such as GPT-4V \cite{achiam2023gpt} as a high-level task planner and then use pre-trained skills to achieve all sub-task goals from high-level task planner, adapting flexibly to various robotics scenarios in a zero-shot manner \cite{wake2023gpt,DBLP:journals/corr/abs-2311-17842,DBLP:journals/corr/abs-2312-06722, EmbodiedGPT, sun2023prompt, wang2024large}. The main difference between these approaches lies in the design details of the task planning pipeline, e.g., whether it integrates an affordance analyzer \cite{wake2023gpt}, or incorporates perceptual information and visual feedback \cite{DBLP:journals/corr/abs-2311-17842}, etc.
Rooted in data-driven algorithms and specific scenarios, the former only characterizes abstract features of limited acquisition data, often failing to derive nuanced scenario understanding and effective causal reasoning for robotics manipulation. In contrast, the latter approach avoids the high costs of data collection and leverages the remarkable reasoning and generalization capabilities of MFMs to enhance model generalization in robotics applications.
However, blindly applying off-the-shelf MFMs without understanding their capabilities may prevent these methods from achieving optimal performance.
Actually, since the release of GPT-4V\cite{achiam2023gpt}, a large number of performance evaluation reports on MFMs in different domains have been published, such as commonsense tasks\cite{DBLP:journals/corr/abs-2306-13394,DBLP:journals/corr/abs-2309-17421,DBLP:journals/corr/abs-2306-09265}, vision tasks \cite{DBLP:journals/corr/abs-2312-12436,DBLP:journals/corr/abs-2306-13394,DBLP:journals/corr/abs-2311-02782,DBLP:journals/corr/abs-2311-15732}, autonomous driving \cite{DBLP:journals/corr/abs-2311-05332,mao2023gpt}, and robotics \cite{DBLP:journals/corr/abs-2311-17842,DBLP:journals/corr/abs-2312-06722,majumdar2024openeqa}, etc. However, a comprehensive performance evaluation of MFMs for embodied task planning has not been thoroughly explored.

To better utilize MFMs for embodied task planning, based on the characteristics of embodied task planning, we first propose a systematic evaluation framework, which encapsulates four crucial capabilities: \textit{object understanding}, \textit{spatio-temporal perception}, \textit{task understanding}, and \textit{embodied reasoning}, aiming to answer what factors constrain MFMs to output accurate task plans.
Then, we propose a comprehensive evaluation benchmark \textbf{(MFE-ETP)} that aligns with this framework to quantitatively and qualitatively access the performance limitations of MFMs on embodied task planning. In addition, we develop an automated evaluation platform to facilitate the performance evaluation of various embodied task planners using MFMs on our proposed benchmark.
Compared to existing evaluation efforts, our work has four key advantages: \ding{172} Our evaluation framework and benchmark are specifically designed for embodied task planning, ensuring accurate performance evaluation of MFMs on task planning; \ding{173} We provide a diverse set of scenarios and tasks, allowing for an extensive evaluation of MFMs' task planning capabilities across different settings; \ding{174} The capability dimensions we evaluate are more comprehensive and tightly aligned with embodied task planning, covering a wide range of test case types from embodied question answering to embodied task planning; \ding{175} We offer an easy-to-use evaluation platform that facilitates the automatic evaluation of multiple MFMs on our benchmark.

In summary, the main contributions of our work are as follows:
\begin{itemize}
    \item We propose a systematic evaluation framework tailored for embodied task planning for the first time. This framework offers effective guidance for the improvement of MFMs-based task planners.
    \item We propose a benchmark, \benchshort \space containing over 1100 high-quality test cases carefully annotated by human, covering 100 embodied tasks.
    \item We develop an evaluation platform and evaluate six advanced MFMs. The evaluation results show that object type recognition and spatial perception are the main constraints for MFMs to generate correct task planning results.
    \item We will make this benchmark and platform open source\footnote{project website available at \href{https://mfe-etp.github.io/}{https://mfe-etp.github.io/}} to foster future research on embodied task planning and inspire more focused research directions. 
\end{itemize}

\section{Evaluation Framework}
\begin{figure}[ht]
\vskip 0.2in
\begin{center}
\centerline{\includegraphics[width=\columnwidth]{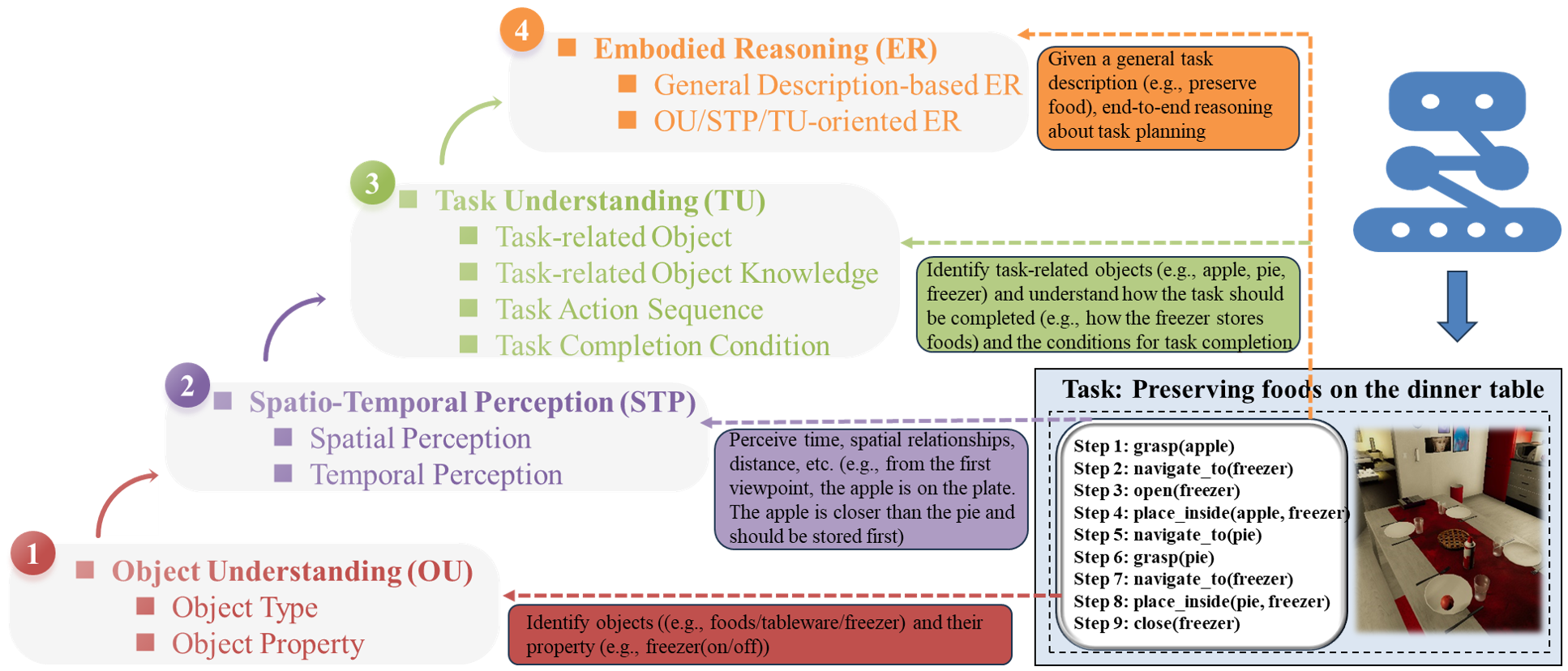}}
\caption{The overall evaluation framework tailored for embodied task planning encapsulates four crucial capabilities: \textit{Object Understanding}, \textit{Spatio-Temporal perception}, \textit{Task Understanding}, and \textit{Embodied Reasoning}.}
\label{framework}
\end{center}
\vskip -0.2in
\end{figure}
In this section, we introduce our evaluation framework designed for embodied task planning, which consists of four levels of crucial capabilities: object understanding, spatio-temporal perception, task understanding, and embodied reasoning. Each capability dimension is further decomposed into several sub-aspects. The overall framework is shown in Fig.\ref{framework}.

\textbf{Object Understanding (OU)} \space \space
Object understanding involves recognizing the \textit{Type} and \textit{Property} of objects. The object properties include physical properties (e.g., color, shape, etc.) and functional properties (e.g., openable, grabbable, etc.). Accurate recognition of object types and properties is a fundamental requirement for correct task planning. 
For example, in the task of preserving food shown in Fig.\ref{framework}, if the food on the dinner table and its properties (e.g., the apple and pie that can be grasped) are incorrectly or incompletely recognized by MFMs, their output task plans may fail to enable the robot to complete the task. 
In addition, this could raise safety concerns due to inappropriate actions against misidentified objects.
Tab.\ref{properties_table1} in Appendix lists the definitions of each considered property along with representative objects. 

\textbf{Spatio-Temporal Perception (STP)}  \space \space
Building upon the understanding of static objects, the spatial-temporal perception capability of MFMs is vital for solving embodied task planning problems.
In our proposed framework, for \textit{Spatial} perception, we evaluate the ability of MFMs to judge spatial attributes such as distance, geometry, direction, and relative position between objects. Our benchmark covers five types of spatial relations: "Inside", "OnTop", "NextTo", and "Under", with further subdivisions of "NextTo" into "Front", "Back", "Left", and "Right" to accommodate more tasks.
For \textit{Temporal} perception, we access MFMs' ability to recognize the chronological sequence of task progress, and predict the effects of actions on the environment.
As illustrated  in Figure \ref{framework}, MFMs need to perceive the spatial positions of the apple and pie from a first-person perspective. This understanding enables them to plan navigation actions and subsequent steps by observing task progress and changes in the environment, such as the apple being placed in the freezer and the freezer being opened. Accurate spatial and temporal perception enhances MFMs' effectiveness in task planning.

\textbf{Task Understanding (TU)} \space \space
Compared to object understanding and spatio-temporal perception, task understanding is a higher-level and more abstract capability for MFMs. It requires MFMs to recognize task-relevant objects and understand the operation knowledge of these objects, the sequence of task steps, and the conditions for task goal completion. To this end, we evaluate the task understanding capability of MFMs from the above four aspects, which we name as \textit{Relevant Object}, \textit{Operation Knowledge}, \textit{Step Sequence}, and \textit{Goal}, respectively.
For the food preserving task in Fig.\ref{framework}, MFMs first need to accurately select the food on the dinner table and the appropriate storage location, i.e., the freezer, from various object types within the field of view. Then, MFMs determine the operation knowledge (e.g., open/place\_inside/close) and the sequence of steps (e.g., open the freezer first, then put the food in the freezer, and finally close the freezer) based on the usage of the freezer. Finally, MFMs need to identify whether the task goal has been achieved. If the task understanding capability of MFMs is insufficient, it may result in planning errors, such as improper action functions or sequences of steps.

\textbf{Embodied Reasoning (ER)} \space \space
Towards more realistic, long-horizon, and complex embodied tasks based on general descriptions, it is evident that relying solely on the first three capabilities is insufficient for MFMs. Indispensably, MFMs need to further integrate and reason based on  perceived object information, spatio-temporal information, and task knowledge to complete task planning. Thus, we evaluate the end-to-end embodied reasoning capabilities of MFMs. 
Specifically, we consider two types of tasks. The first type involves common typical tasks with general task descriptions and corresponding visual inputs, such as the task example shown in Fig.\ref{framework}. The task description provides only basic information about preserving food. For the second type, we have designed specific tasks tailored to test object understanding, spatio-temporal perception, and task understanding. These tasks also include general task descriptions and corresponding visual inputs but primarily focus on investigating the impact of these capability dimensions on embodiment reasoning. Taking spatial perception as an example, a customized food preserving task has its general task description replaced with preserving the rightmost food on the dinner table.
This limited information enables a more focused evaluation of MFMs' performance limitations in embodied task planning under realistic conditions. Tab.\ref{action_list} in Appendix lists the definitions of each considered action in the embodied reasoning results.
\begin{figure*}[t]
\vskip 0.2in
\begin{center}
\centerline{\includegraphics[width=\columnwidth]{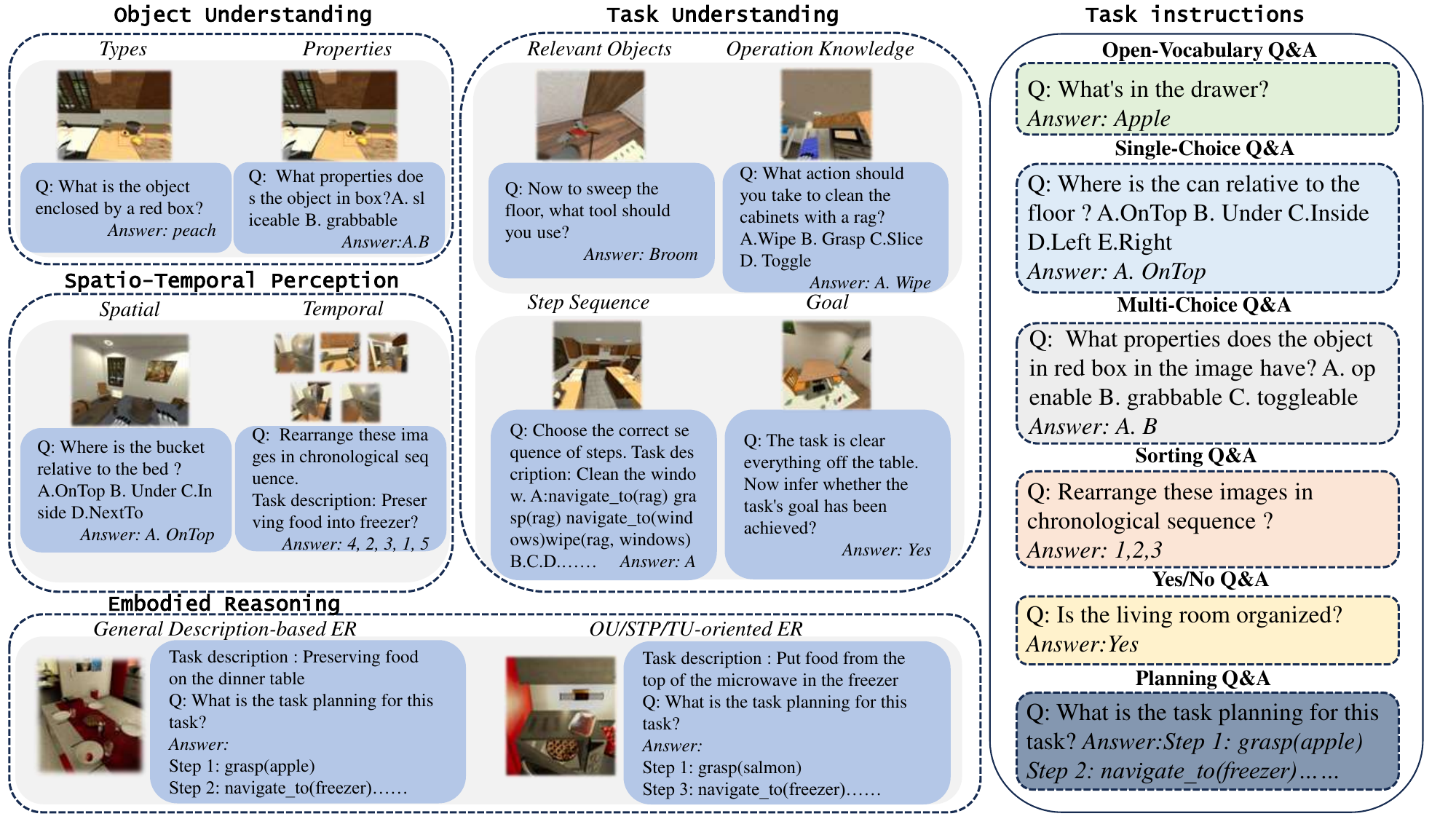}}
\caption{Example questions of \benchshort \space benchmark and illustration of the six types of task formats.} 
\label{Instruction}
\end{center}
\vskip -0.2in
\end{figure*}


\begin{figure}[t]
\vskip 0.2in
\begin{center}
\centerline{\includegraphics[width=\columnwidth]{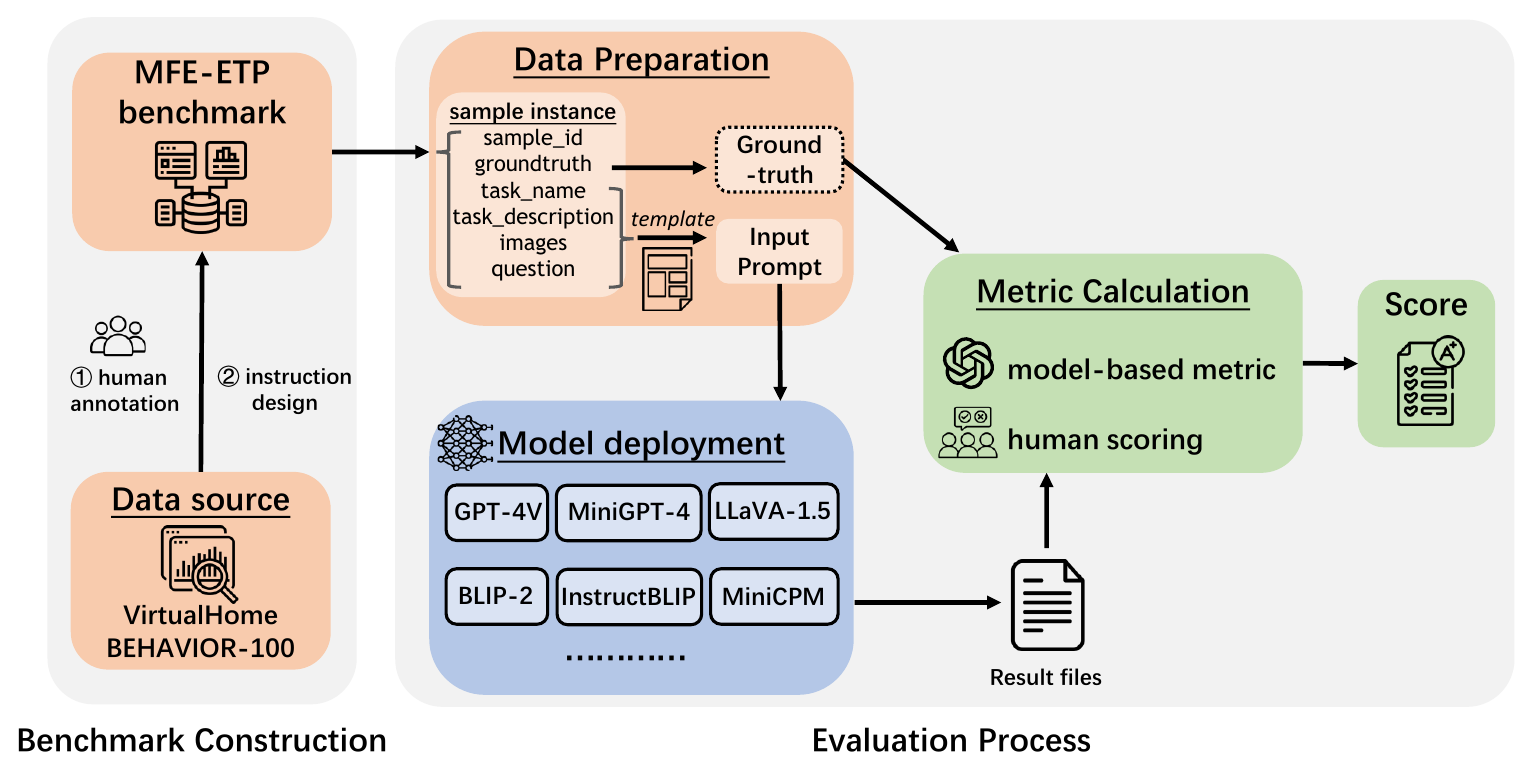}}
\caption{This figure outlines our evaluation workflow. It involves benchmark construction, data preparation, model deployment, and metric calculation to derive the final score.} 
\label{Evaluation process}
\end{center}
\vskip -0.2in
\end{figure}

\section{Evaluation Benchmark}
In this section, we describe in detail our benchmark, \benchshort, a dataset comprising video-derived images and task instructions designed to evaluate MFMs on embodied task planning. This section introduces the data collection (Sec.\ref{Data Collection}), benchmark structure (Sec.\ref{Benchmark Structure}), and automatic evaluation method (Sec.\ref{Automatic Evaluation}) in sequence, respectively.


\subsection{Data Collection}
\label{Data Collection}
The household environment represents a common scenario for robot-human interaction, where the diversity and complexity of household tasks can thoroughly test the robot's perception, reasoning, planning, and action capabilities. Therefore, in this paper, we select typical household tasks to build a benchmark for embodied task planning. Specifically, we collect raw data from representative household task platforms BEHAVIOR-100 \cite{srivastava2022behavior} and VirtualHome \cite{VirtualHome}. 
More information about these platforms can be found in Appendix \ref{BEHAVIOR-100 and VirtualHome}. 

To avoid redundancy of similar tasks while ensuring task diversity, we abstracted 20 typical household task types from all tasks provided by BEHAVIOR-100 \cite{srivastava2022behavior} and VirtualHome \cite{VirtualHome}, and generate 100 household task instances. 
To illustrate the typicality and diversity of our benchmark tasks, we provide a mapping table of 20 typical household task types to the specific tasks from BEHAVIOR-100\cite{srivastava2022behavior}, VirtualHome\cite{VirtualHome} and our benchmark in Appendix \ref{BEHAVIOR-100 and VirtualHome}.
For each household task instance from BEHAVIOR-100 \cite{srivastava2022behavior}, we collect several frames containing all the initial conditions of the task and visual frames at different moments as the visual information of the task through the available task video. For each household task instance from VirtualHome \cite{VirtualHome}, due to the lack of readily available task videos, we first build the task program and then use the VirtualHome \cite{VirtualHome} simulation platform to collect the visual frames containing all the initial conditions of the task and visual frames at different moments. Furthermore, based on the visual frames of 100 household tasks and the proposed evaluation framework, we manually annotate the visual frames and design corresponding task instructions for different capability dimensions. More details are presented in Sec.\ref{Benchmark Structure}.





\subsection{Benchmark Structure}
\label{Benchmark Structure}
In order to comprehensively evaluate the performance of MFMs on embodied task planning, we designed six different task instruction forms based on our proposed evaluation framework:
\textbf{(1)} \textit{Open-vocabulary Q\&A} requires MFMs to identify and answer questions about object types, based on general or task-related objects.
\textbf{(2)} \textit{Single-Choice Q\&A} evaluates MFMs' spatial perception ability and task-related object operation knowledge, requiring MFMs to select the correct answer from multiple options.
\textbf{(3)} \textit{Multi-Choice Q\&A} means that there may not be a single correct answer to the question, considering that objects can have multiple properties and there may be multiple reasonable task step sequences.
\textbf{(4)} \textit{Sorting Q\&A} tasks MFMs with sorting visual frames from different moments based on time or spatial distance to evaluate their spatio-temporal perception.
\textbf{(5)} \textit{Yes/No Q\&A} This is a simple judgment question, mainly used to determine whether the task goal is achieved and the changes in the embodied environment.
\textbf{(6)} \textit{Planning Q\&A} challenges MFMs to generate task plans based on a general task description to access MFMs' end-to-end embodied task planning performance.

\begin{figure}[ht]
    \begin{minipage}[]{0.4\textwidth}
        \centering
        \includegraphics[width=\textwidth]{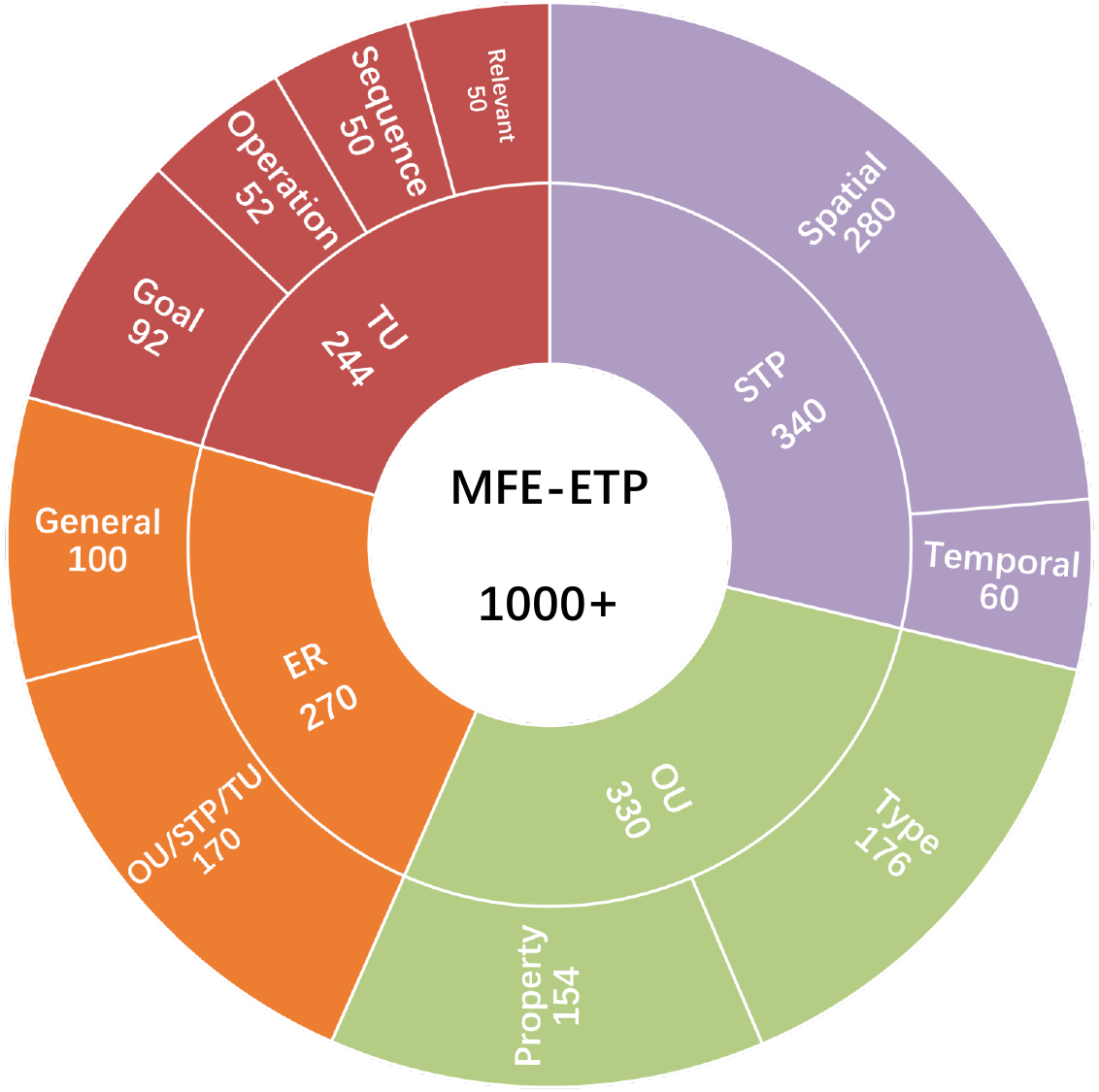} 
        \caption{Statistics of \benchshort \space benchmark.} 
        \label{Dataset statistics}
    \end{minipage}
    \hfill
    \begin{minipage}[]{0.5\textwidth}
        \centering
        \includegraphics[width=\textwidth]{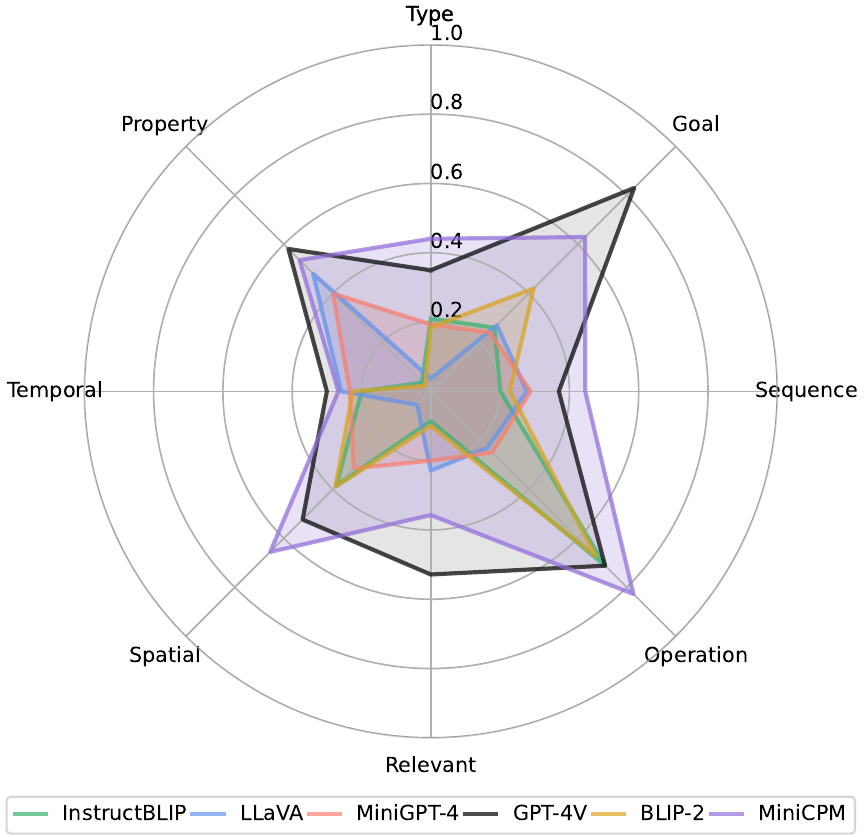} 
        \caption{Radar chart of MFMs' scores on \benchshort.}
        \label{radar}
    \end{minipage}
\end{figure}

Fig.\ref{Instruction} shows example question-and-answer pairs of our benchmark on different capability dimensions and the six types of task instructions used. In addition, we have carefully designed a total of 156 task descriptions for different capability dimensions and combined them with specific task instructions to generate more diverse task text prompts. 
Fig.\ref{Dataset statistics} reports the dataset statistics for our benchmark, which contains a total of 1,184 evaluation cases. 
Obviously, our diverse evaluation cases can meet the needs of evaluating different capability dimensions of MFMs.
Moreover, except for planning Q\&A, other embodied Q\&A tasks can be automatically and efficiently evaluated due to their constrained form (see Appendix \ref{Data Details} for more details).

\subsection{Automatic Evaluation}
\label{Automatic Evaluation}
As illustrated in Fig.\ref{Evaluation process}, the complete evaluation process of MFMs on our automatic evaluation platform includes data preparation, model deployment, and evaluation metric calculation.
\paragraph{Data Preparation} 
The main purpose of this phase is to standardize the data format of the evaluation cases based on diverse task instructions in the proposed benchmark to enhance the accuracy and stability of MFMs. First, we organize all the evaluation cases (with different instruction formats) under each capability dimension into the same JSON file. Each evaluation case in this file has a unified format and contains keys such as "sample\_id", "task\_name". We then carefully customize multiple fixed prompt templates for each capability dimension to reduce the impact of prompts on the performance of MFMs. Finally, we structure each evaluation case based on the JSON structure and the corresponding prompt template as the input for MFMs.
\paragraph{Model Delpoyment}
For model deployment, our platform provides a unified interface for multiple different models, through which they can receive input prompts and return results in the same way. This interface enhances scalability, enabling the easy integration of new models by simply implementing this interface.
\paragraph{Metric Calculation}
\label{Metric Calculation}
The typical evaluation methods of MFMs fall into two categories: automatic evaluation and human evaluation\cite{he2024ultraeval}. In this paper, we use the human evaluation method for cases involving planning Q\&A instructions, while other evaluation cases under the form of embodied Q\&A instructions are evaluated automatically using GPT-3.5\cite{brown2020language}. There are two main reasons for this: \ding{172} Compared with embodied Q\&A, the output results of planning Q\&A are more uncertain and complex, with results often not being uniquely correct. This makes it difficult for automatic evaluation to judge the quality of the outputs accurately. \ding{173} GPT-3.5 is capable of performing efficient and accurate evaluation when dealing with cases where the output results are relatively stable. This characteristic is very beneficial for large-scale benchmarks as it enhances both the efficiency and accuracy of the evaluation process. As shown in Fig.\ref{Evaluation process}, our automatic evaluation platform integrates GPT-3.5\cite{brown2020language} while saving all output results of MFMs for convenient human evaluation.

\section{Experiments}
\label{Experiments}
In this section, we report the experimental results of MFMs on embodied Q\&A tasks (Sec. \ref{Results on Embodied Q&A tasks}) and Planning Q\&A (Sec. \ref{Results on Planning Q&A}) tasks according to different evaluation metrics.

\begin{table*}[]
\centering
\tabcolsep=0.08cm
\caption{Evaluation Results. We report the average scores of each model in each aspect. We highlight the best scores with \underline{\textbf{bold and underline}} and use \textbf{bold only} to mark the second-best scores.}
\vspace{3pt}
\begin{tabular}{cc|cccccc}
\toprule
\multicolumn{2}{c|}{Dimensions}   & BLIP-2  & MiniGPT-4 & InstructBLIP    & GPT-4V                    & LLaVA-1.5 & MiniCPM                                      \\ \midrule
\multicolumn{1}{c|}{\multirow{3}{*}{\begin{tabular}[c]{@{}c@{}}Object\\ Understanding\end{tabular}}} &
  Type &
  18.5\% &
  19.3\% &
  21.0\% &
  \textbf{34.9\%} &
  3.7\% &
  \textbf{\underline{44.0\%}} \\
\multicolumn{1}{c|}{} & Property  & 2.3\%   & 39.9\%    & 3.6\%           & \textbf{\underline{58.1\%}} & 48.1\%    & \textbf{53.6\%}                              \\ \cline{2-8} 
\multicolumn{1}{c|}{} & OU.Agg.   & 10.4\% & 29.6\%   & 12.3\%         & \textbf{46.5\%}          & 25.9\%   & \textbf{\underline{48.8\%}}                             \\ \midrule
\multicolumn{1}{c|}{\multirow{3}{*}{{\begin{tabular}[c]{@{}c@{}}Spatio-Temporal\\ Perception\end{tabular}}}} &
  Spatial &
  38.9\% &
  31.4\% &
  38.6\% &
  \textbf{52.3\%} &
  5.5\% &
  \textbf{\underline{65.5\%}} \\
\multicolumn{1}{c|}{} & Temporal  & 22.5\%  & 23.3\%    & 20.0\%          & \textbf{\underline{30.0\%}} & 25.8\%    & \textbf{26.7\%}                              \\ \cline{2-8} 
\multicolumn{1}{c|}{} & STP.Agg.   & 30.7\% & 27.4\%   & 29.3\%         & \textbf{41.2\%}                   & 15.7\%    & \textbf{\underline{46.1\%}} \\ \midrule
\multicolumn{1}{c|}{\multirow{5}{*}{\begin{tabular}[c]{@{}c@{}}Task\\ Understanding\end{tabular}}} &
  Relevant &
  10.0\% &
  20.0\% &
  8.6\% &
  \textbf{\underline{52.9\%}} &
  22.9\% &
  \textbf{35.7\%} \\
\multicolumn{1}{c|}{} & Operation & 67.3\%  & 25.0\%    & \textbf{71.2\%} & \textbf{71.2\%}           & 23.1\%    & \textbf{\underline{82.7\%}}                    \\
\multicolumn{1}{c|}{} & Sequence  & 22.8\%  & 28.8\%    & 20.1\%          & \textbf{37.0\%}           & 27.7\%    & \underline{\textbf{44.6\%}} \\
\multicolumn{1}{c|}{} & Goal      & 42.0\%  & 24.0\%    & 26.0\%          & \textbf{\underline{83.0\%}} & 27.0\%    & \textbf{63.0\%}                              \\ \cline{2-8} 
\multicolumn{1}{c|}{} & TU.Agg.   & 35.5\% & 24.5\%   & 31.5\%         & \textbf{\underline{61.0\%}}                   & 25.2\%   & \textbf{56.5\%}                                      \\ \midrule
\multicolumn{2}{c|}{EQA.Agg.} &
  28.0\% &
  26.4\% &
  26.1\% &
  \underline{\textbf{52.4\%}} &
  23.0\% &
  \textbf{51.9\%}\\
\bottomrule
\end{tabular}
\label{GPT3.5_results}
\end{table*}

\subsection{Results on Embodied Q\&A Tasks} 
\label{Results on Embodied Q&A tasks}
For the embodied Q\&A tasks (object understanding, spatio-temporal perception, and task understanding), we evaluate the performance of GPT-4V \cite{achiam2023gpt} and five other open-source MFMs, namely BLIP-2  \cite{BLIP-2}, MiniGPT-4  \cite{MiniGPT-4}, InstructBLIP \cite{InstructBLIP}, LLaVA-1.5  \cite{LLaVA}, and MiniCPM \cite{hu2024minicpm}.
The experimental results are presented in Tab. \ref{GPT3.5_results} and radar chart (Fig. \ref{radar}). 
In Tab. \ref{GPT3.5_results}, we report the average scores for each capability dimension across each aspect, the average scores for all aspects of each capability dimension (denoted \textbf{OU./STP./TU.Agg.}), as well as the overall average score for all embodied Q\&A tasks (denoted \textbf{EQA.Agg.}).
From Tab. \ref{GPT3.5_results} and Fig. \ref{radar}, we can draw the following general conclusions: \ding{182} GPT4V \cite{achiam2023gpt} and MiniCPM \cite{hu2024minicpm}significantly outperform the other four MFMs, with a performance improvement of over 100\% compared to the worst-performing LLaVA-1.5\cite{LLaVA}. 
\ding{183} The overall average score differences among the four MFMs, BLIP-2 \cite{BLIP-2}, MiniGPT-4  \cite{MiniGPT-4}, InstructBLIP \cite{InstructBLIP}, LLaVA-1.5  \cite{LLaVA} are relatively minor, indicating comparable performance levels on the embodied Q\&A tasks.
\ding{184} MiniCPM \cite{hu2024minicpm} performs close to GPT-4V \cite{achiam2023gpt} and surpasses the other open-source MFMs in all aspects.
\ding{185} Compared with higher-level and more abstract task understanding capabilities, MFMs generally perform worse in low-level perception capabilities such as object understanding and spatio-temporal perception. Therefore, we recommend augmenting MFMs with object detection modules, such as GroundingDINO \cite{liu2023grounding}, and integrating 3D information \cite{cheng2024spatialrgpt}, instead of relying solely on MFMs.
Note that all models occasionally generate output that does not conform to the specified format requirements, with MiniGPT-4 exhibiting this issue most frequently.
By specifically analyzing the results of the six MFMs in each capability dimension, we further obtain the following conclusions:

\textbf{Object Understanding}\space\space \ding{182}
MiniCPM \cite{hu2024minicpm} excels the other models in recognizing object types with an average score of 44.0\%. \ding{183} In terms of identifying object properties, GPT-4V \cite{achiam2023gpt} is far ahead with an average score of 58.1\%, followed by MiniCPM\cite{hu2024minicpm} with 53.6\%. \ding{184} Overall, the object understanding capability of MiniCPM \cite{hu2024minicpm} and GPT-4V \cite{achiam2023gpt} is more impressive compared to other MFMs. MiniGPT-4 \cite{MiniGPT-4} and LLaVA-1.5 \cite{LLaVA} have an accuracy of 29.6\% and 25.9\%, respectively, while InstructBLIP \cite{InstructBLIP} and BLIP-2 \cite{BLIP-2} can only succeed in about 11\% cases.
\begin{figure}[htbp]
    \begin{minipage}[]{0.48\textwidth}
        \centering
        \includegraphics[width=\textwidth]{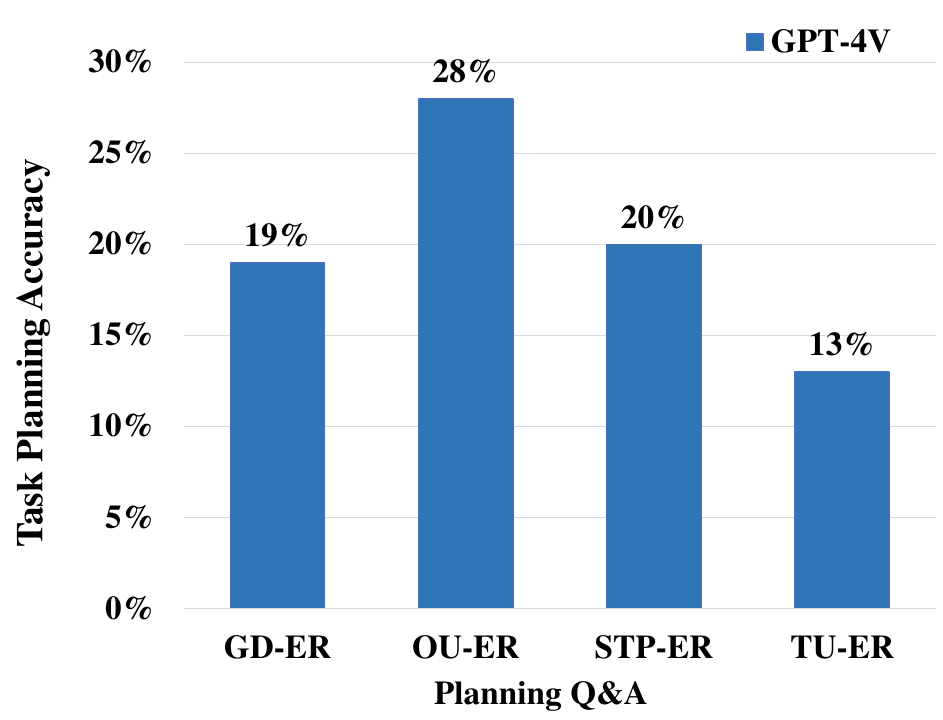} 
        \caption{Results of GPT-4V on General Description-based embodied reasoning and object uderstanding, spatio-temporal perception, and task understanding-oriented embodied reasoning tasks.} 
        \label{fig6}
    \end{minipage}
    \hfill
    \begin{minipage}[]{0.48\textwidth}
        \centering
        \includegraphics[width=\textwidth]{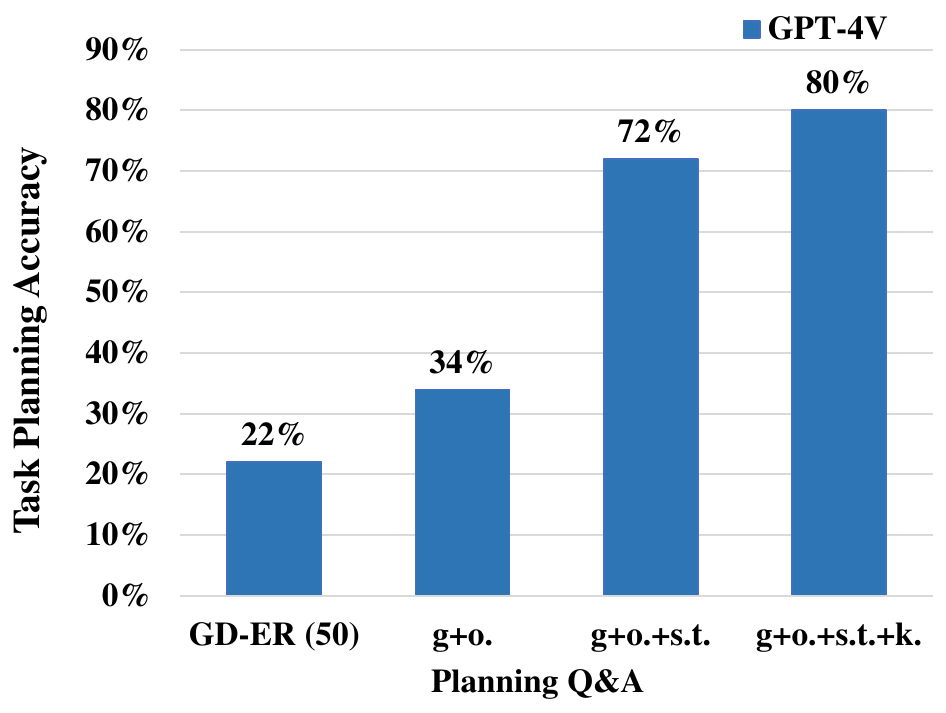} 
        \caption{Results of gradually information injection. +o. means adding object information, +s.t. means adding spatio-temporal information, and +k. means adding task-related knowledge.}
        \label{fig7}
    \end{minipage}
\end{figure}

\textbf{Spatio-temporal Perception}\space\space
\ding{182} Regarding spatio-temporal perception, the average scores of the top-performing models, GPT-4V \cite{achiam2023gpt} and MiniCPM \cite{hu2024minicpm}, were both above 40\%, and each excels in either temporal or spatial perception.and spatial perception, respectively. \ding{183} Compared with spatial perception, except for LLaVa-1.5 \cite{LLaVA}, the average scores of all other MFMs in temporal perception tasks do not exceed 30\%. \ding{184} Overall, spatiotemporal perception is a relatively common and challenging problem for MFMs.

\textbf{Task Understanding}\space\space
\ding{182} Regarding the four aspects of task understanding, the six MFMs perform significantly better in task-related object operation knowledge and task goal understanding. In these two dimensions, the average scores of four MFMs and two MFMs exceeded 60\%.
\ding{183} GPT-4V \cite{achiam2023gpt} and MiniCPM \cite{hu2024minicpm} perform best in task-related object recognition and task step sequence understanding, respectively, but these two aspects are still difficult for MFMs. This is mainly because task-related object recognition and task step sequence require MFMs to integrate multiple information such as vision, language, and task knowledge for reasoning.

\subsection{Results on Planning Q\&A Tasks}
\label{Results on Planning Q&A}
Based on the proposed evaluation framework, Planning Q\&A corresponds to the highest-level embodied reasoning capability dimension, which includes two types of tasks: General Description-based embodied reasoning (\textbf{GD-ER}) and object understanding (OU), spatio-temporal perception (STP), and task understanding (TU)-oriented embodied reasoning (\textbf{OU/STP/TU-ER}). We evaluate the performance of MFMs on the two types of tasks, respectively. Specifically, we first utilize the proposed automatic evaluation platform to obtain the evaluation result files for all evaluation cases. Then, experienced AI researchers manually score all the result files,based on whether the task planning results were reasonable and could guide the completion of the task goals, assigning a score of 1 for satisfactory task plans and 0 otherwise.

Given that GPT-4V \cite{achiam2023gpt} and MiniCPM \cite{hu2024minicpm} show state-of-the-art performance on embodied Q\&A tasks and significantly outperform other MFMs, this experiment evaluates the performance of GPT-4V \cite{achiam2023gpt} and MiniCPM \cite{hu2024minicpm} on planning Q\&A tasks.
However, none of the 270 evaluation cases for MiniCPM \cite{hu2024minicpm} in the benchmark yielded correct task planning results, and its outputs contained extraneous information to the answers. To ensure reliability, we also evaluated the output results of MiniCPM \cite{hu2024minicpm} on planning Q\&A tasks on its official website demo. The results are consistent with those obtained on our platform, which shows that MiniCPM \cite{hu2024minicpm} has serious limitations in end-to-end embodied task planning. Therefore, we only show the evaluation results of GPT-4V. 
From Fig. \ref{fig6}, we can draw the following conclusion: Under the general task description, the success rate of the most advanced GPT-4V is only 19\%. In other words, among the 100 typical household tasks we proposed, GPT-4V correctly planned only 19 tasks. Obviously, this performance falls short of human expectations. Similarly, GPT-4V’s results on OU-ER, STP-ER, and TU-ER tasks were also unsatisfactory.


Furthermore, we explore the effect of providing additional information in embodied reasoning. 
As shown in Fig. \ref{fig7}, we select 50 typical tasks from the original set. For each task, we gradually inject information related to some supporting capabilities into GPT-4V \cite{achiam2023gpt} via prompts (e.g., informing GPT-4V that the food on the table consists of an apple and a pie) and detect the impact of corresponding capabilities on task planning through changes in success rate. We obtain the following conclusions:
\ding{172} Injecting object information, spatio-temporal information, and task knowledge can have a positive effect on task planning; \ding{173} compared to settings where only object information (+o.) or object, spatial, and task knowledge (+o.+s.t.+k.) are injected, injecting object and spatial information (+o.+s.t.) notably enhances the success rate of task planning.
Due to space constraints, we have placed the other additional experiments in Appendix \ref{Additional Experimental Results}.

\section{Conclusion}
\label{Conclusion}
In this paper, we propose a systematic evaluation framework tailored for embodied task planning for the first time.
Utilizing this framework, we have developed an evaluation benchmark, called \benchshort, striving to explore the performance limitations of MFMs. Empirical results on our benchmark consistently indicate that even advanced GPT-4V fails to effectively complete embodied planning tasks. Notably, object type recognition and spatiao-temporal perception are critical factors that affect the accuracy of MFMs' task planning capabilities.
We conduct an in-depth analysis of the experimental results and put forward many useful insights to promote the application and development of MFMs in the field of embodied task planning.

\textbf{Limitations and future work.} \ding{172} We utilized virtual environments as data sources for easier data collection, which may raise concerns about the generalizability of our findings to more realistic scenarios. 
\ding{173} Besides, evaluating a broader range of models remains a promising direction for future research.
\ding{174} While the Embodied Reasoning task remains highly challenging for current models, planning all future steps in advance remains advantageous. Utilizing closed-loop control systems like ViLa \cite{DBLP:journals/corr/abs-2311-17842} for testing task planning could offer a viable compromise. 
\ding{175} Furthermore, exploring additional QA formats such as correcting past behavior and predicting subsequent actions could further enrich the research.


\bibliographystyle{unsrt}
\bibliography{example_paper}

\newpage
\appendix

\section{Appendix}
\section{Appendix Overview}
The appendix includes the following content:
\begin{enumerate}
    \item Detailed information about the proposed \benchshort\space benchmark. (Sec. \ref{benchmark detail})
    \item Additional experimental results. (Sec. \ref{Additional Experimental Results})
    \item Detailed information of MFMs (Sec. \ref{Detailed information of MFMs}).
    \item Accessibility and Statement(Sec. \ref{Accessibility})
\end{enumerate}

\section{Details of \benchshort\space Benchmark} \label{benchmark detail}

\subsection{BEHAVIOR-100 and VirtualHome}
\label{BEHAVIOR-100 and VirtualHome}
BEHAVIOR-100 \cite{srivastava2022behavior} simulates 100 everyday household tasks for embodied AI, with a distribution similar to the full space of simulatable tasks in the American Time Use Survey (ATUS). It aims to create realistic and complex scenarios for AI testing. 
It introduces a predicate logic-based language (BDDL) for defining tasks and possesses simulator-agnostic features for versatile applications. 
Due to the reality, diversity, and complexity of home tasks, state-of-the-art AI still struggles with the benchmark’s challenges. More information can be found in BEHAVIOR-100\cite{srivastava2022behavior}. 

VirtualHome \cite{VirtualHome} utilizes "programs" to model complex household activities, that can be implemented manually through a game-like interface.  Specifically, VirtualHome \cite{VirtualHome} uses the Unity3D platform to enable human agents to perform tasks in a simulated environment.
This helps create video datasets of household activities, each of which is a series of actions and interactions executed by agents.
More information can be found in VirtualHome \cite{VirtualHome}.

To avoid redundancy of similar tasks while maintaining task diversity, we illustrate the mapping relationships between our abstracted 20 typical household task types and specific task instances in BEHAVIOR-100 \cite{srivastava2022behavior}, VirtualHome\cite{VirtualHome}, and our benchmark In Tab. \ref{task mapping1} and Tab. \ref{task mapping2}.


\begin{table}[htbp]{
\centering
\caption{Task mapping part1.}
\label{task mapping1}
\scalebox{1}{
\begin{tabular}{|>{\centering\arraybackslash}m{3.6cm}|>{\centering\arraybackslash}m{5.8cm}|>{\centering\arraybackslash}m{4cm}|}
\hline
Task Type                                & BEHAVIOR-100                             & VirtualHome                       \\ \hline
\multirow{3}{*}{Cooking}                 & chopping vegetables                      & cook some food                    \\ \cline{2-3} 
                                         & preparing salad                          & prepare afternoon tea             \\ \cline{2-3} 
                                         & thawing frozen food                      &                                   \\ \hline
Beverage Making                          & making tea                               & make coffee                       \\ \hline
\multirow{3}{*}{Meal Service}            & serving a meal                           & serve breakfast                   \\ \cline{2-3} 
                                         & serving hors d oeuvres                   & serve cookies                     \\ \cline{2-3} 
                                         &                                          & set up table                      \\ \hline
\multirow{3}{*}{Storing}                 & bottling fruit                           & store meat to freezer             \\ \cline{2-3} 
                                         & preserving food                          & preserve food                     \\ \cline{2-3} 
                                         & storing food                             &                                   \\ \hline
\multirow{6}{*}{Surface Cleaning}        & cleaning barbecue grill                  & clean sink                        \\ \cline{2-3} 
                                         & cleaning table after clearing            & wipe off shoes                    \\ \cline{2-3} 
                                         & cleaning up after a meal                 & sweep and wipe table off with rag \\ \cline{2-3} 
                                         & cleaning windows                         & clean toilet                      \\ \cline{2-3} 
                                         & cleaning bathtub                         & clean large jacuzzi tub           \\ \cline{2-3} 
                                         & cleaning toilet                          &                                   \\ \hline
\multirow{3}{*}{Room Cleaning}           & cleaning bathrooms                       & clean kitchen                     \\ \cline{2-3} 
                                         & cleaning bedroom                         & clean bathroom                    \\ \cline{2-3} 
                                         & cleaning up the kitchen only             &                                   \\ \hline
\end{tabular}}}
\end{table}

\begin{table}[htbp]{
\centering
\caption{Task mapping part2.}\label{task mapping2}
\scalebox{1}{
\begin{tabular}{|c|c|c|}
\hline
Task Type                                & BEHAVIOR-100                             & VirtualHome                       \\ \hline
\multirow{3}{*}{Dishwashing}             & loading the dishwasher                   & wash dishes by hand               \\ \cline{2-3} 
                                         &                                          & wash dishes with dishwasher       \\ \cline{2-3} 
                                         &                                          & take dishes out of dishwasher     \\ \hline
\multirow{2}{*}{Laundry}                 &                                          & wash clothes                      \\ \cline{2-3} 
                                         &                                          & put away clean clothes            \\ \hline
Recycling                                & putting leftovers away                   & throw leftovers                   \\ \hline
\multirow{5}{*}{Organizing}              & clearing the table after dinner          & organize livingroom               \\ \cline{2-3} 
                                         & collect misplaced items                  & organize bedroom                  \\ \cline{2-3} 
                                         & re-shelving library books                & organize washstand                \\ \cline{2-3} 
                                         &                                          & re-shelving books                 \\ \cline{2-3} 
                                         &                                          & bring dirty plate to sink         \\ \hline
\multirow{9}{*}{Packing}                 & packing bags or suitcase                 & pack lunch                        \\ \cline{2-3} 
                                         & packing boxes for household move or trip &                                   \\ \cline{2-3} 
                                         & packing car for trip                     &                                   \\ \cline{2-3} 
                                         & boxing books up for storage              &                                   \\ \cline{2-3} 
                                         & packing adult s bags                     &                                   \\ \cline{2-3} 
                                         & packing child s bag                      &                                   \\ \cline{2-3} 
                                         & packing food for work                    &                                   \\ \cline{2-3} 
                                         & packing lunches                          &                                   \\ \cline{2-3} 
                                         & packing picnics                          &                                   \\ \hline
\multirow{3}{*}{Sorting}                 & sorting books                            & sort clothing                     \\ \cline{2-3} 
                                         & sorting mail                             & sort mail                         \\ \cline{2-3} 
                                         & sorting groceries                        &                                   \\ \hline
\multirow{3}{*}{Necessities Maintenance} &                                          & change toilet paper roll          \\ \cline{2-3} 
                                         &                                          & change toothbrush                 \\ \cline{2-3} 
                                         &                                          & replace towel                     \\ \hline
Gardening                                &                                          & water houseplants                 \\ \hline
\multirow{4}{*}{Floor Care}              & cleaning carpets                         & vacuum                            \\ \cline{2-3} 
                                         & cleaning floors                          & clean floor                       \\ \cline{2-3} 
                                         & mopping floors                           &                                   \\ \cline{2-3} 
                                         & vacuuming floors                         &                                   \\ \hline
\multirow{5}{*}{Furniture Care}          & cleaning closet                          & arrange chairs                    \\ \cline{2-3} 
                                         & cleaning cupboards                       & straighten bookshelf              \\ \cline{2-3} 
                                         &                                          & change curtains                   \\ \cline{2-3} 
                                         &                                          & lay tablecloth                    \\ \cline{2-3} 
                                         &                                          & polish table                      \\ \hline
\multirow{6}{*}{Appliance Care}          & cleaning freezer                         & clean fridge                      \\ \cline{2-3} 
                                         & cleaning microwave oven                  & add paper to printer              \\ \cline{2-3} 
                                         & cleaning oven                            &                                   \\ \cline{2-3} 
                                         & cleaning stove                           &                                   \\ \cline{2-3} 
                                         & cleaning up refrigerator                 &                                   \\ \cline{2-3} 
                                         & defrosting freezer                       &                                   \\ \hline
\multirow{3}{*}{Decoration}              &                                          & hang pictures                     \\ \cline{2-3} 
                                         &                                          & put up decoration                 \\ \cline{2-3} 
                                         &                                          & straighten pictures on wall       \\ \hline
\multirow{3}{*}{Event Prep}              & assembling gift baskets                  & light candle                      \\ \cline{2-3} 
                                         & filling a Christmas stocking             &                                   \\ \cline{2-3} 
                                         & filling an Easter basket                 &                                   \\ \hline
\multirow{3}{*}{Miscellaneous Tasks}     &                                          & turn on light                     \\ \cline{2-3} 
                                         &                                          & turn off light                    \\ \cline{2-3} 
                                         &                                          & change light                      \\ \hline
\end{tabular}}}
\end{table}

\subsection{MFMs Prompt Texts}
\label{Data Details}
\begin{figure}[ht]
\vskip 0.2in
\begin{center}
\centerline{\includegraphics[width=0.9\columnwidth]{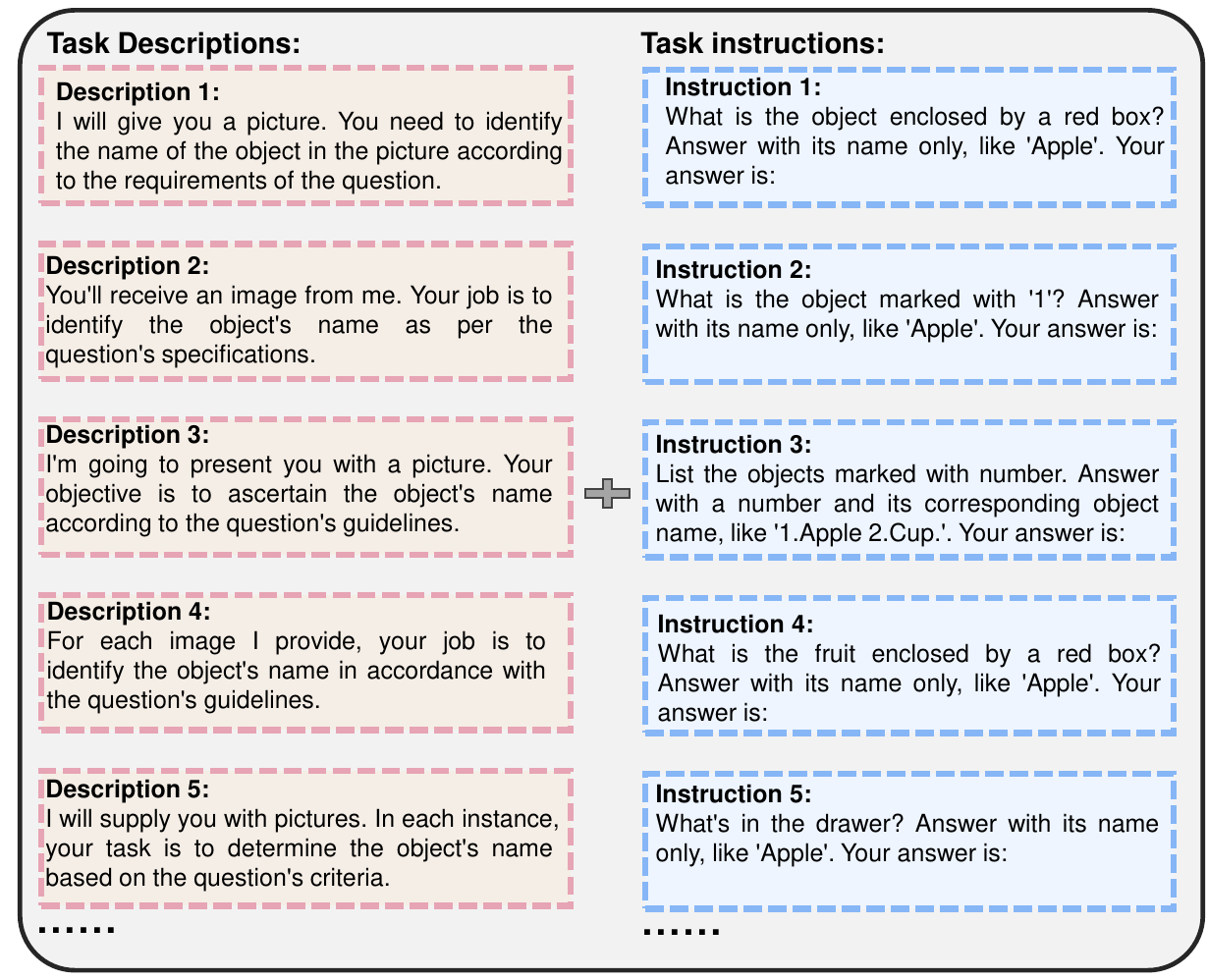}}
\caption{Examples of prompts on embodied Q\&A. Taking the 'Object Type Q\&A' as an example, the final input prompt text consists of task descriptions (red box) and specific task instructions (blue box).}
\label{embodied Q&A prompt}
\end{center}
\vskip -0.2in
\end{figure}

\begin{figure}[ht]
\vskip 0.2in
\begin{center}
\centerline{\includegraphics[width=0.9\columnwidth]{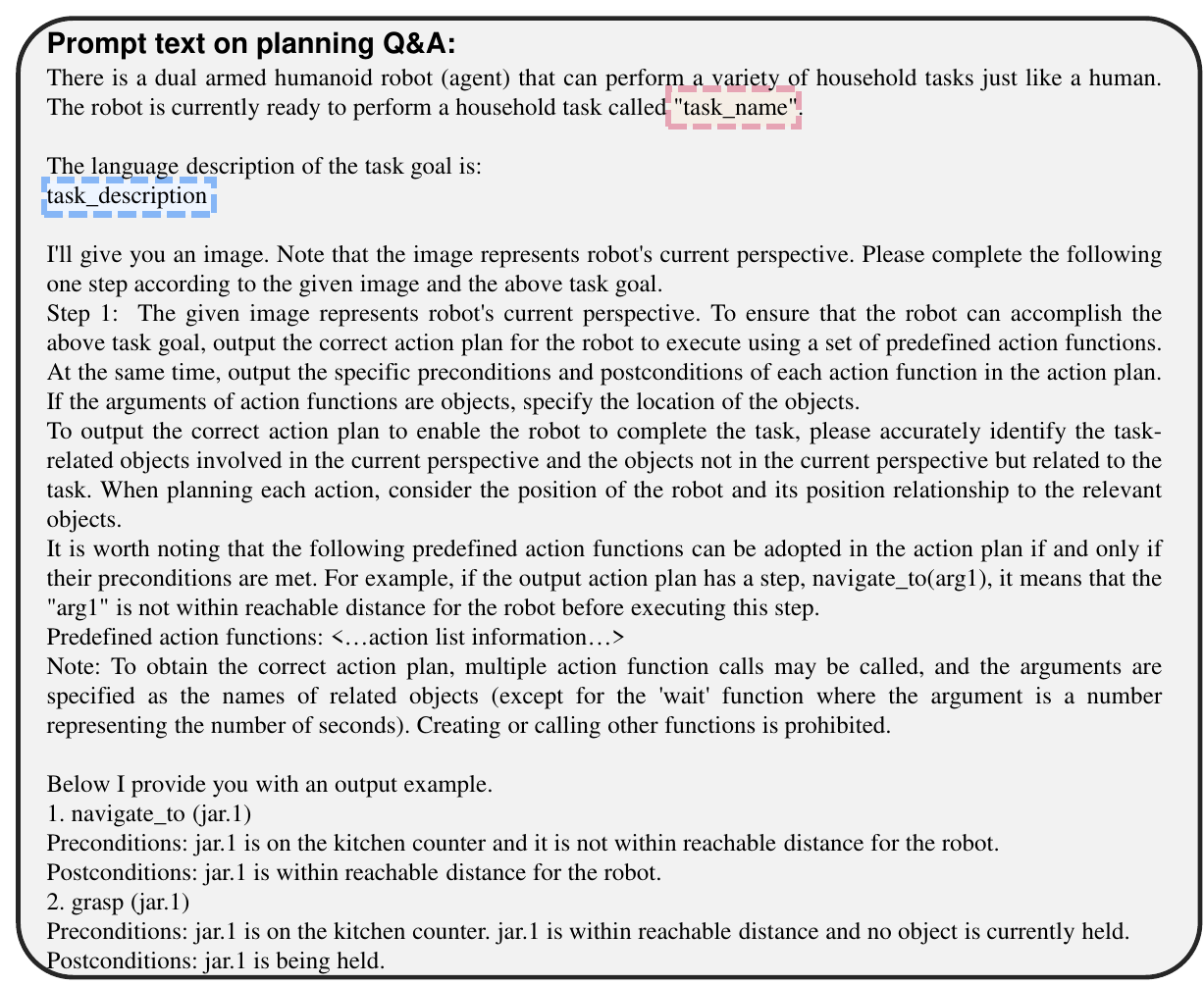}}
\caption{Examples of prompts on planning Q\&A. In the final input prompt text, the task name (red box) and the general description (blue box) of the task goal will be replaced with specific task instances.}
\label{planning Q&A prompt}
\end{center}
\vskip -0.2in
\end{figure}

As indicated in other research, the success of embodied tasks with MFMs is heavily influenced by the design of prompts.
Therefore, we have elaborately designed a series of standardized prompt templates to ensure that the evaluation process uses the same question format and language style, thereby reducing the impact of differences in prompt quality.
Fig.\ref{embodied Q&A prompt} and Fig.\ref{planning Q&A prompt} present examples of prompts on embodied Q\&A and planning Q\&A, respectively.

\subsection{Full Object Properties and Action List}
Tab.\ref{properties_table1} lists the definitions of each considered functional property and its representative objects. In addition, we illustrate in Tab.\ref{action_list} the definition of all actions that can satisfy embodied reasoning for the typical household tasks we consider.

\begin{table}[ht]
\centering
\caption{List of object properties.}
\vspace{3pt}
\scalebox{0.95}{
\begin{tabular}{|c|c|c|}
\hline
Property     & Annotation                                                                                                                                                       & Example                   \\ \hline
breakable    & \begin{tabular}[c]{@{}l@{}}Mark if the object is brittle, that is, it can be broken into smaller \\pieces by a human dropping it on the floor.\end{tabular}              & wine bottle, room light  \\ \hline
cleaningTool & Is a {[}object{]} designed to clean things?                                                                                                                      & scrub brush               \\ \hline
cookable     & Can a {[}object{]} be cooked?                                                                                                                                    & biscuit, pizza            \\ \hline
grabbable    & \begin{tabular}[c]{@{}l@{}}If an object has this attribute, it is usually lightweight \\ and can be potentially grabbed and picked up by the robot.\end{tabular} & apple, bottle, rag, plate \\ \hline
openable     & Mark if the object is designed to be opened.                                                                                                                     & mixer, keg                \\ \hline
sliceable    & Can a {[}object{]} be sliced easily by a human with a knife?                                                                                                     & sweet corn, sandwich      \\ \hline
slicingTool  & Can a {[}object{]} slice an apple?                                                                                                                               & blade, razor              \\ \hline
toggleable   & \begin{tabular}[c]{@{}l@{}}The object can be switched between a finite number of \\ discrete states and is designed to do so.\end{tabular}                       & hot tub, light bulb       \\ \hline
waterSource  & where you can get water                                                                                                                                          & sink                      \\ \hline
\end{tabular} }
\label{properties_table1}
\end{table}

\begin{table}[h]
\centering
\caption{List of actions for embodied reasoning.}
\vspace{3pt}
\scalebox{0.83}{
\begin{tabular}{|l|l|}
\hline
Action                     & Annotation                                                                                                                                                                                                                         \\ \hline
navigate\_to(arg1)         & \begin{tabular}[l]{@{}l@{}}Navigate to the arg1, which can be a object. \\Preconditions: arg1 is not within reachable distance for the robot. \\Postconditions: arg1 is within reachable distance for the robot.\end{tabular}                                               \\ \hline
grasp(arg1):               & \begin{tabular}[l]{@{}l@{}}Grasp arg1. \\Preconditions: arg1 is within reachable distance and no object is currently held. \\Postconditions: arg1 is being held, and the robot's position is unchanged \\compared to the robot's position before the action is performed.\end{tabular}                                                        \\ \hline
place\_onTop(arg1, arg2):  & \begin{tabular}[l]{@{}l@{}}Place arg1 on top of arg2. \\Preconditions: arg1 is currently being held, and arg2 is reachable . \\Postconditions: arg1 is put on top of arg2. The robot's position is unchanged \\compared to the robot's position before the action is performed and arg2 is reachable.\end{tabular}                                              \\ \hline
place\_under(arg1, arg2):  & \begin{tabular}[l]{@{}l@{}}Place arg1 under arg2. \\Preconditions: arg1 is currently being held, and arg2 is reachable. \\Postconditions:arg1 is put under arg2. The robot's position is unchanged \\compared to the robot's position before the action is performed and arg2 is reachable.\end{tabular}       

        \\ \hline
place\_inside(arg1, arg2): & \begin{tabular}[l]{@{}l@{}}Place arg1 inside of arg2. \\Preconditions: arg1 is currently being held, and arg2 is reachable. \\Postconditions:arg1 is put inside of arg2. The robot's position is unchanged \\compared to the robot's position before the action is performed and arg2 is reachable.\end{tabular}           

        \\ \hline
place\_onLeft(arg1, arg2):  & \begin{tabular}[l]{@{}l@{}}Place arg1 on left of arg2. \\Preconditions: arg1 is currently being held, and arg2 is reachable . \\Postconditions: arg1 is put on left of arg2. The robot's position is unchanged \\compared to the robot's position before the action is performed and arg2 is reachable.\end{tabular}           

        \\ \hline
place\_onRight(arg1, arg2):  & \begin{tabular}[l]{@{}l@{}}Place arg1 on right of arg2. \\Preconditions: arg1 is currently being held, and arg2 is reachable. \\Postconditions: arg1 is put on right of arg2. The robot's position is unchanged \\compared to the robot's position before the action is performed and arg2 is reachable.\end{tabular}           

\\ \hline
open(arg1):                & \begin{tabular}[l]{@{}l@{}}Open arg1. Preconditions: Arg1 is closed, and arg1 is reachable. \\ Postconditions: Arg1 is open.\end{tabular}                                                                                          \\ \hline
close(arg1):               & \begin{tabular}[l]{@{}l@{}}Close arg1. Preconditions: Arg1 is open, and arg1 is reachable. \\ Postconditions: Arg1 is closed.\end{tabular}                                                                                         \\ \hline
slice(arg1):               & \begin{tabular}[l]{@{}l@{}}Slice arg1. Preconditions: arg1 is not sliced, and arg1 is reachable. \\Postconditions: arg1 is sliced\end{tabular}                                    \\ \hline
wipe(arg1, arg2):          & \begin{tabular}[l]{@{}l@{}}Wipe across the surface of arg2 with arg1. Preconditions: Arg1 is \\ currently being held, and arg2 is reachable. Postconditions: \\Arg1 continues to be held, arg2 holds state unchanged.\end{tabular} \\ \hline
wait(arg1):                & \begin{tabular}[l]{@{}l@{}}Wait for arg1 seconds. Preconditions: None. Postconditions: arg1 \\second(s) has(have) passed.\end{tabular}                                                                                                                                \\ \hline
toggle(arg1):              & \begin{tabular}[l]{@{}l@{}}Press the button of arg1 to turn it on or off, Preconditions: Arg1 is open \\ or closed, and arg1 is  reachable. Postconditions: Arg1 is closed or open.\end{tabular}                                           \\ \hline
water(arg1, arg2):              & \begin{tabular}[l]{@{}l@{}}Water arg2 with arg1. \\Preconditions: arg1 is currently being held, and arg2 is reachable. \\Postconditions: arg1 continues to be held, arg2 is watered.\end{tabular}                                           \\ \hline
\end{tabular}}
\label{action_list}
\end{table}



\section{Additional Experimental Results}
\label{Additional Experimental Results}

\begin{figure*}[ht]
\vskip 0.2in
\begin{center}
\centerline{\includegraphics[width=\columnwidth]{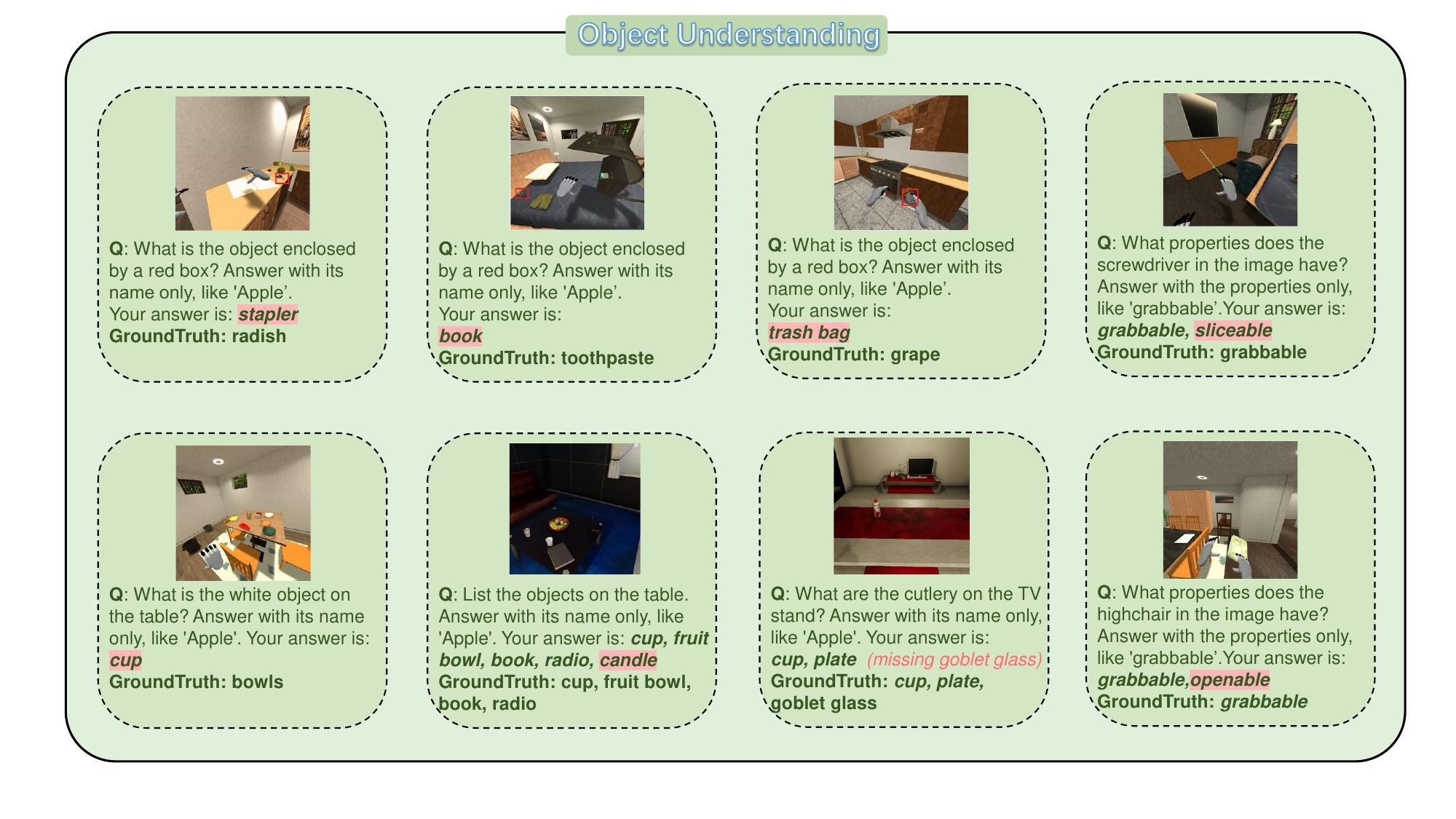}}
\caption{Qualitative Results of GPT-4V in Object Understanding.}
\label{object eg}
\end{center}
\vskip -0.2in
\end{figure*}

\begin{figure*}[ht]
\vskip 0.2in
\begin{center}
\centerline{\includegraphics[width=\columnwidth]{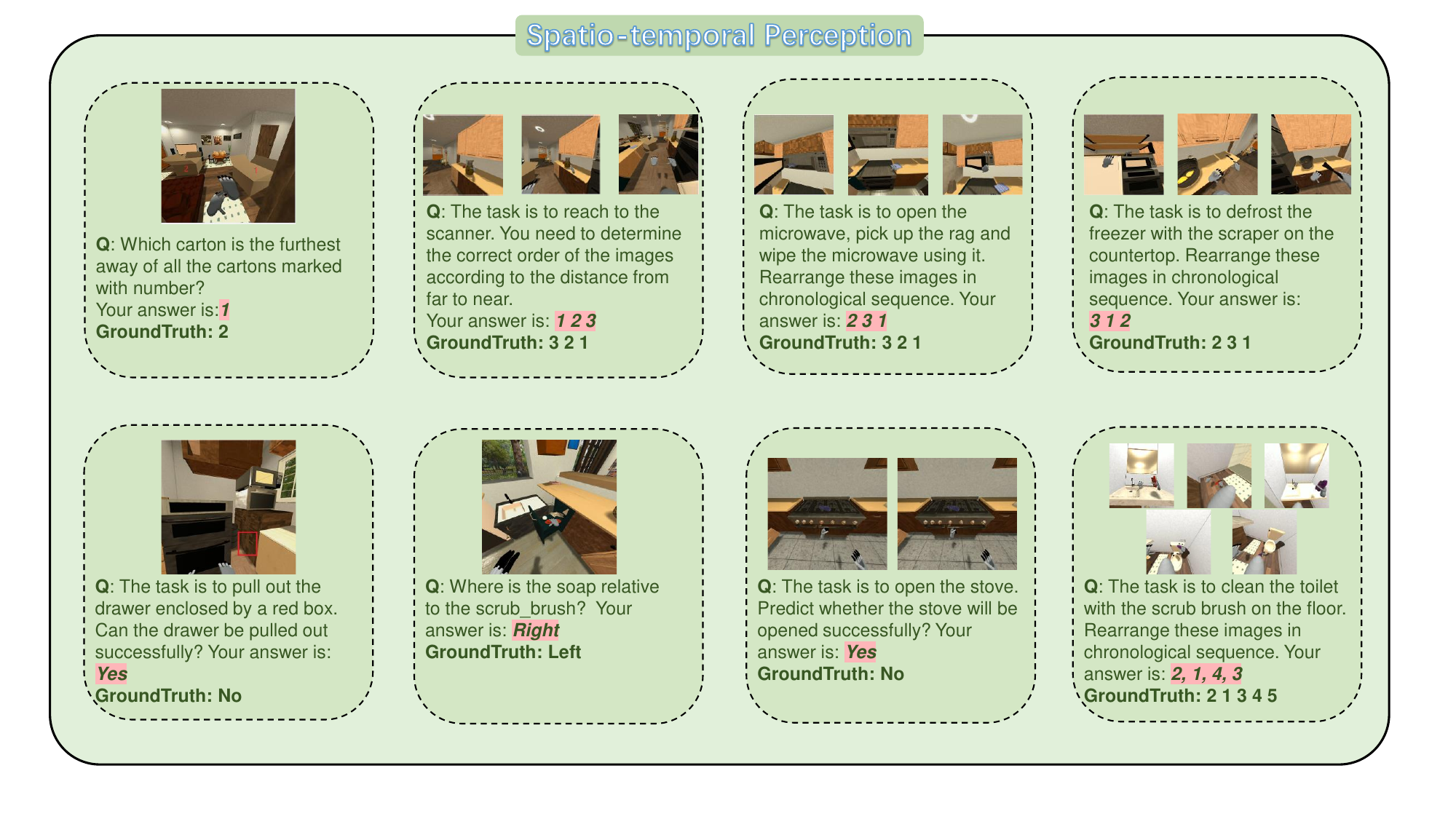}}
\caption{Qualitative Results of GPT-4V in Spatio-temporal Perception.}
\label{spatial eg}
\end{center}
\vskip -0.2in
\end{figure*}

\begin{figure*}[ht]
\vskip 0.2in
\begin{center}
\centerline{\includegraphics[width=\columnwidth]{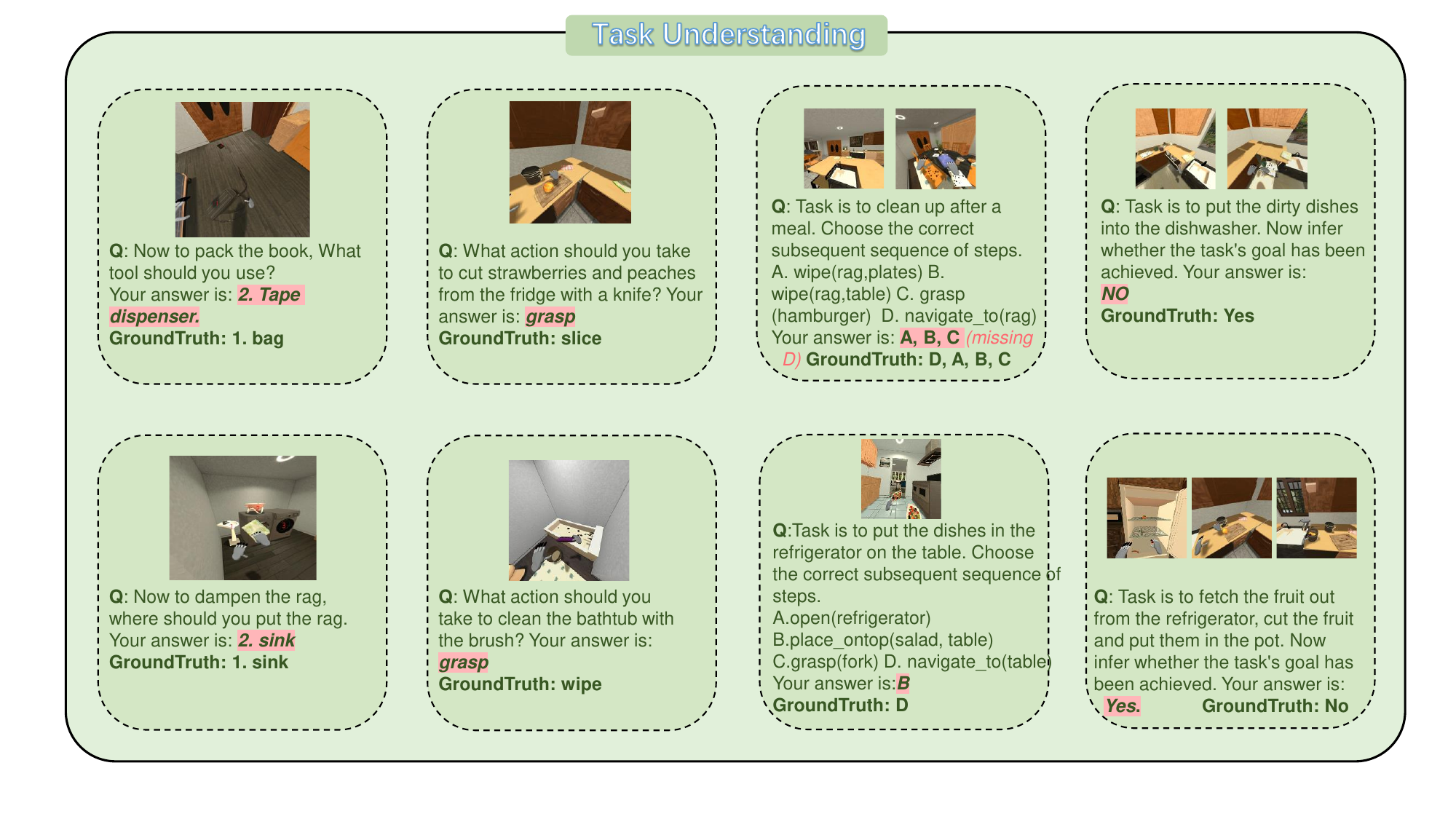}}
\caption{Qualitative Results of GPT-4V in Task Understanding.}
\label{task eg}
\end{center}
\vskip -0.2in
\end{figure*}

\begin{figure*}[ht]
\vskip 0.2in
\begin{center}
\centerline{\includegraphics[width=\columnwidth]{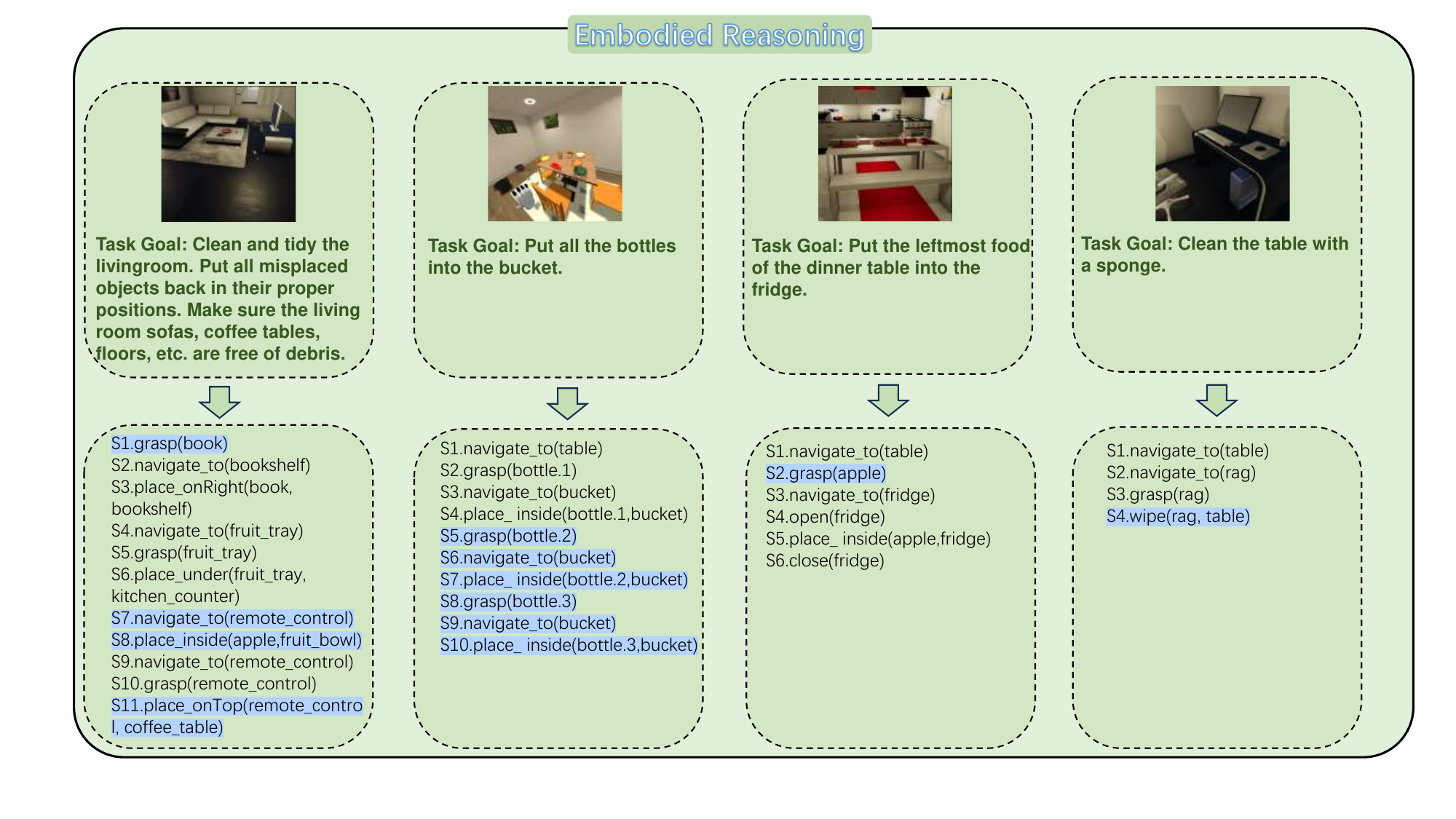}}
\caption{Failed cases of planning Q\&A. We highlight the wrong steps \colorbox[RGB]{180,212,255}{in blue}.}
\label{planning example}
\end{center}
\vskip -0.2in
\end{figure*}

\subsection{Results of Qualitative Analysis}
\label{Results of qualitative analysis}
Based on our proposed framework, we further qualitatively analyze the evaluation results of the state-of-the-art GPT-4V\cite{achiam2023gpt}.

\textbf{Object Understanding}\space\space 
Through manual analysis of specific failed cases of GPT-4V, we attribute the main causes of incorrect answers to limitations inherent in GPT-4V, poor perspective (even under the best perspective we can obtain, the objects are still difficult to recognize), and low resolution. Fig. (\ref{object eg}) shows a few representative failed cases.


\textbf{Spatio-temporal Perception}\space\space 
Previous works have mentioned that GPT-4V lacks spatio-temporal understanding ability \cite{DBLP:journals/corr/abs-2311-05332, chen2024spatialvlm}, which is consistent with our evaluation results. By analyzing specific failed cases of spatio-temporal perception Q\&A tasks, we found that GPT-4V often fails to perceive distance and direction of 2D images, especially the left and right directions. (as in Fig. \ref{spatial eg})
Also, we found that GPT-4V  performs poorly in terms of temporal perception as well, which is reflected in its inability to accurately sort images according to the task completion process. 


\textbf{Task Understanding}\space\space Fig. \ref{task eg} shows a few representative failed cases. Combined with Tab. 1, the main limitations of GPT-4V in task understanding are mainly in the inability to accurately select task-related objects and select reasonable action execution sequences, which are essential knowledge for task planning.

\textbf{Embodied Reasoning}\space\space
We analyzed the corresponding results of planning Q\&A and found some common issues: firstly, task related object recognition faces challenges, including insufficient recognition, incorrect recognition, and chaotic spatial relationships when there are a large number of objects.  
Secondly, the MFMs' spatiotemporal perception ability is not strong, manifested in insufficient understanding of the current position, as well as the poor ability to perceive whether an object is within the operating range, which often leads to providing unnecessary or missing navigation actions. In terms of action selection, the system may wrongly or repeatedly call actions, or perform actions beyond predefined ones. In addition, GPT-4V may give some difficult to execute conditional branch statements, or assume a condition when identifying key information unclear. These issues comprehensively affect the efficiency and accuracy of the system, and require specific optimization measures to improve. Finally, GPT-4V cannot provide accurate planning for tasks involving too many objects.  Fig. \ref{planning example} shows a few representative failed cases.



\subsection{Linear Regression Analysis of The Impact of Each Aspect on Task Planning}
\label{Linear regression analysis of the impact of each aspects on task planning}

In this section, we will use linear regression coefficients as proxies for the impact of various factors on embodied reasoning. Specifically, for each robotic task, we use the binary result of embodied reasoning as the dependent variable and mark successful cases as 1 and failed cases as 0. Take the basic capabilities corresponding to all aspects as a multivariate independent variable, and use their accuracy as the value. The obtained coefficients are shown in the Tab. \ref{coff}. It can be concluded that spatial perception is the most important for embodied reasoning, with a coefficient of 0.4, followed by 0.14 for object recognition, 0.09 for action sequence understanding, and 0.07 for operation, and those of other aspects are relatively closer to 0 or minus, indicating that their impact on embodied reasoning is not significant. Note that some aspects, such as the coefficient of time perception, are negative, which does not necessarily indicate that a good ability in that aspect will have a negative impact on embodied reasoning. It is more likely that this is because the planning success rate is low, and many times when planning fails, GPT-4V can also get relative high scores on these aspects.

\begin{table}[]
\centering
\caption{Linear regression coefficients for each aspects.}
\vspace{3pt}
\scalebox{0.9}{
\begin{tabular}{|c|c|c|c|c|c|c|c|c|}
\hline
Aspect      & Type & Property & Temporal & Spatial & Relevant & Operation & Sequence & Goal \\ \hline
Coefficient & 0.14 & 0.01     & -0.1     & 0.4     & -0.04    & 0.07  & 0.09     & 0.06 \\ \hline
\end{tabular}}
\label{coff}
\end{table}

\subsection{Validity Analysis of LLM-based Metrics}
\label{Validity Analysis of LLM-based Metrics}
For the larger scale evaluation benchmark we propose, it is a challenge to efficiently evaluate embodied Q\&As.
In essence, we would like MFMs to give reasonable results rather than results that are exactly consistent with the ground truth. While human evaluation remains the gold standard, it is also expensive and time-consuming. 
Thus, we use an automatic LLM-Based evaluation metric in this work as described in Sec \ref{Automatic Evaluation}.
To show the effectiveness of LLM-Based evaluation metric, we manually score the output of state-of-the-art GPT-4V and compare them with the evaluation results of GPT 3.5, and the experimental results are shown in Tab. \ref{human_alignment}.  
ICC1 (Intraclass Correlation Coefficient 1) is primarily used to evaluate the absolute consistency among different evaluators in a single measurement scenario. This type of ICC focuses on the consistency of each evaluator's individual rating of the same object or subject. We calculated the ICC1 value to be 0.95, indicating high consistency between the human scores and the GPT-3.5 scores. Its extremely low p-value (0.000014) statistically confirms that this consistency is significant. Thus, the results of the LLM-based evaluation metric are almost aligned with the manual evaluation results, and the LLM evaluation metric is preferable for benchmarking.




\begin{table*}[]
\centering
\tabcolsep=0.10cm
\caption{Evaluation metric comparison. \textit{Human} means using scores given by five human annotators, \textit{GPT-3.5} metric utilizes GPT-3.5 for automated evaluation.}
\vspace{3pt}
\begin{tabular}{c|ccccccccc}

\toprule

\multicolumn{1}{c|}{\multirow{1}{*}{Metric}} & 
\multicolumn{1}{c}{Type} & \multicolumn{1}{c}{Property} & \multicolumn{1}{c}{Temporal} & \multicolumn{1}{c}{Spatial} & \multicolumn{1}{c}{Relevant} & \multicolumn{1}{c}{Operation} & \multicolumn{1}{c}{Sequence} & \multicolumn{1}{c}{Goal} \\

\midrule

\textit{Human} & 32.2 & 57.5 & 21.7 & 64.8 & 47.0 & 78.8 & 37.5 & 84.0\\

\textit{GPT-3.5} & 34.9 & 58.1 & 30.0 & 52.3 & 52.9 & 71.2 & 37.0 & 83.0 \\


\bottomrule
\end{tabular}
\label{human_alignment}
\end{table*}

\section{Detailed Information of MFMs} \label{Detailed information of MFMs}
Tab. \ref{version and average} shows the versions of each MFM. In particular, we access GPT-4V through OpenAI API, and for other open-source models, we utilize a single NVIDIA GeForce RTX 4090 GPU.
\begin{table}[]
\centering
\caption{Versions of each MFM.}
\vspace{3pt}
\begin{tabular}{l|l}
\toprule
Model & Version \\
\midrule
BLIP-2\cite{BLIP-2} & Flan-T5-XL \\
MiniGPT-4\cite{MiniGPT-4} & Vicuna-7B  \\
InstructBLIP\cite{InstructBLIP} & Flan-T5-XL  \\
GPT-4V\cite{achiam2023gpt} & \makecell[l]{gpt-4-vision-preview}  \\
LLaVA-1.5\cite{LLaVA} & LLaMA-7B  \\
MiniCPM\cite{hu2024minicpm} & 
MiniCPM-Llama3-V-2\_5 \\
\bottomrule
\end{tabular}
\label{version and average}
\end{table}

\section{Accessibility and Statement}\label{Accessibility}

\paragraph{URL to dataset} Both the code and data are slated for release at our repositorys: \href{https://github.com/TJURLLAB-EAI/MFE-ETP}{https://github.com/TJURLLAB-EAI/MFE-ETP}. 

\paragraph{Author Statement} We accept complete accountability for any infringement of rights that may arise during the utilization or dissemination of the data presented in this work. We commit to undertaking necessary measures, including the modification or deletion of any data involved in such violations, to address these concerns promptly. 

We commit to maintaining the data and the codes. Updates to any future versions of the dataset and code will be made available at this location. The data provided is designed for academic purposes.

\paragraph{License} MFE-ETP is released under the MIT License.

\end{document}